\documentclass{article} % For LaTeX2e
\usepackage{iclr2025_conference,times}

\usepackage{amsmath,amsfonts,bm}

\def\eqref#1{equation~\ref{#1}}
\def\1{\bm{1}}

\DeclareMathAlphabet{\mathsfit}{\encodingdefault}{\sfdefault}{m}{sl}
\SetMathAlphabet{\mathsfit}{bold}{\encodingdefault}{\sfdefault}{bx}{n}

\usepackage{hyperref}
\usepackage{url}
\usepackage{graphicx}

\title{Improving World Models using\\Deep Supervision with Linear Probes}

\author{%
Andrii Zahorodnii \\
Department of Electrical Engineering and Computer Science, MIT\\
Department of Brain and Cognitive Sciences \& McGovern Institute for Brain Research, MIT\\
Cambridge, MA, USA \\
\texttt{zaho@csail.mit.edu}
}

\iclrfinalcopy

\begin{document}

\maketitle
\begin{abstract}
Developing effective world models is crucial for creating artificial agents that can reason about and navigate complex environments. 
In this paper, we investigate a deep supervision technique for encouraging the development of a world model in a network trained end-to-end to predict the next observation. 
While deep supervision has been widely applied for task-specific learning, our focus is on improving the world models.
Using an experimental environment based on the Flappy Bird game, where the agent receives only LIDAR measurements as observations, we explore the effect of adding a linear probe component to the network's loss function. This additional term encourages the network to encode a subset of the true underlying world features into its hidden state. 
Our experiments demonstrate that this supervision technique improves both training and test performance, enhances training stability, and results in more easily decodable world features -- even for those world features which were not included in the training. Furthermore, we observe a reduced distribution drift in networks trained with the linear probe, particularly during high-variability phases of the game (flying between successive pipe encounters). 
Including the world features loss component roughly corresponded to doubling the model size, suggesting that the linear probe technique is particularly beneficial in compute-limited settings or when aiming to achieve the best performance with smaller models. These findings contribute to our understanding of how to develop more robust and sophisticated world models in artificial agents, paving the way for further advancements in this field.
\end{abstract}

\section{Introduction}
Developing effective world models is a crucial aspect of artificial intelligence, as it can enable agents to make accurate predictions of how the world will unfold, as well as how their actions will influence the world, and plan their actions accordingly \citep{NEURIPS2018_2de5d166}. World models allow agents to reason about the environment and its dynamics. In the context of AI, a "world model" typically refers to a learned approximation of the environment's underlying dynamics, which may include physical laws, object interactions, and agent-environment relationships. By compressing sensory observations into a structured latent space, world models allow agents to reason about their surroundings and adapt to changing conditions. However, learning robust and accurate world models remains a significant challenge, especially in complex and partially observable environments.

% Jaedong
Recent advancements in machine learning have shown promise in learning world models through end-to-end training of predictive recurrent neural networks (RNNs; \citet{hafner2020dreamcontrollearningbehaviors}). These networks are trained to predict the next observation based on the current observation and action, essentially learning a model of the environment's dynamics. However, it is unclear whether such networks can develop implicit world models that capture the true underlying structure and features of the environment.

In this paper, we consider an example environment -- Flappy Bird with LIDAR (Figure~\ref{fig:env}) -- and investigate whether end-to-end trained predictive RNNs develop implicit world models. Crucially, we explore techniques to encourage the emergence of such models. Specifically, we focus on the following research question:
Can an addition of a linear probe to the network's loss function, which encourages the decoding of true world features, improve the network's performance and the quality of the learned world model?
This technique lies within the broader class of deep supervision \citep{pmlr-v38-lee15a, li2022comprehensivereviewdeepsupervision}, but its application to the problem of world model learning remains relatively unexplored.

By investigating these questions and techniques, we aim to contribute to the understanding and development of more effective world models in artificial intelligence, ultimately leading to more capable and adaptable agents.

\subsection{Related Work}
Deep supervision techniques were first introduced by \citep{pmlr-v38-lee15a} and were extensively applied in training vision models \citep{AlBarazanchi2016NovelCA, Ai2017HumanPE, zou2023objectdetection20years}, with prominent applications in the field of medical imaging \citep{zhou2017deepsupervisionpancreaticcyst, lin2021automated, mishra2019ultrasound, LEI2018290}, image super-resolution \cite{7780550, 8434354} and segmentation \citep{9363892, yan2022multiscale}.
For a comprehensive review, see \citet{li2022comprehensivereviewdeepsupervision}. Linear probing of world models is also a technique in mechanistic interpretability \citep{li2024emergentworldrepresentationsexploring, gurnee2024languagemodelsrepresentspace}, although it is usually not used for \emph{training} better world models.

Learning recurrent world models to learn behaviors in the latent space for downstream applications was introduced by \citet{NEURIPS2018_2de5d166}, and advanced by the Dreamer sequence of models \citep{hafner2020dreamcontrollearningbehaviors, hafner2022masteringataridiscreteworld, hafner2024masteringdiversedomainsworld} as well as other techniques for learning world models such as contrastive learning \citep{kipf2020contrastivelearningstructuredworld} and applications of pretrained models of the world to physical settings such as robotics \citep{pmlr-v205-wu23c, firoozi2024foundation}

Representations of continuous variables in recurrent neural networks (RNNs), as well as mechanisms for their storage and evolution are extensively studied in the field of computational neuroscience \citep{khona2022attractor, noorman2024maintaining, NEURIPS2023_361e5112, burak2009accurate, darshan2022learning}, investigating stable attractor mechanisms \citep{cueva2021recurrentneuralnetworkmodels}, as well as periodic or quasi-periodic ones \citep{park2023persistentlearningsignalsworking}. % add rnn variables; the state of the world
\section{Methods}

\begin{figure}[h]
  \centering
  \includegraphics[width=1.0\textwidth]{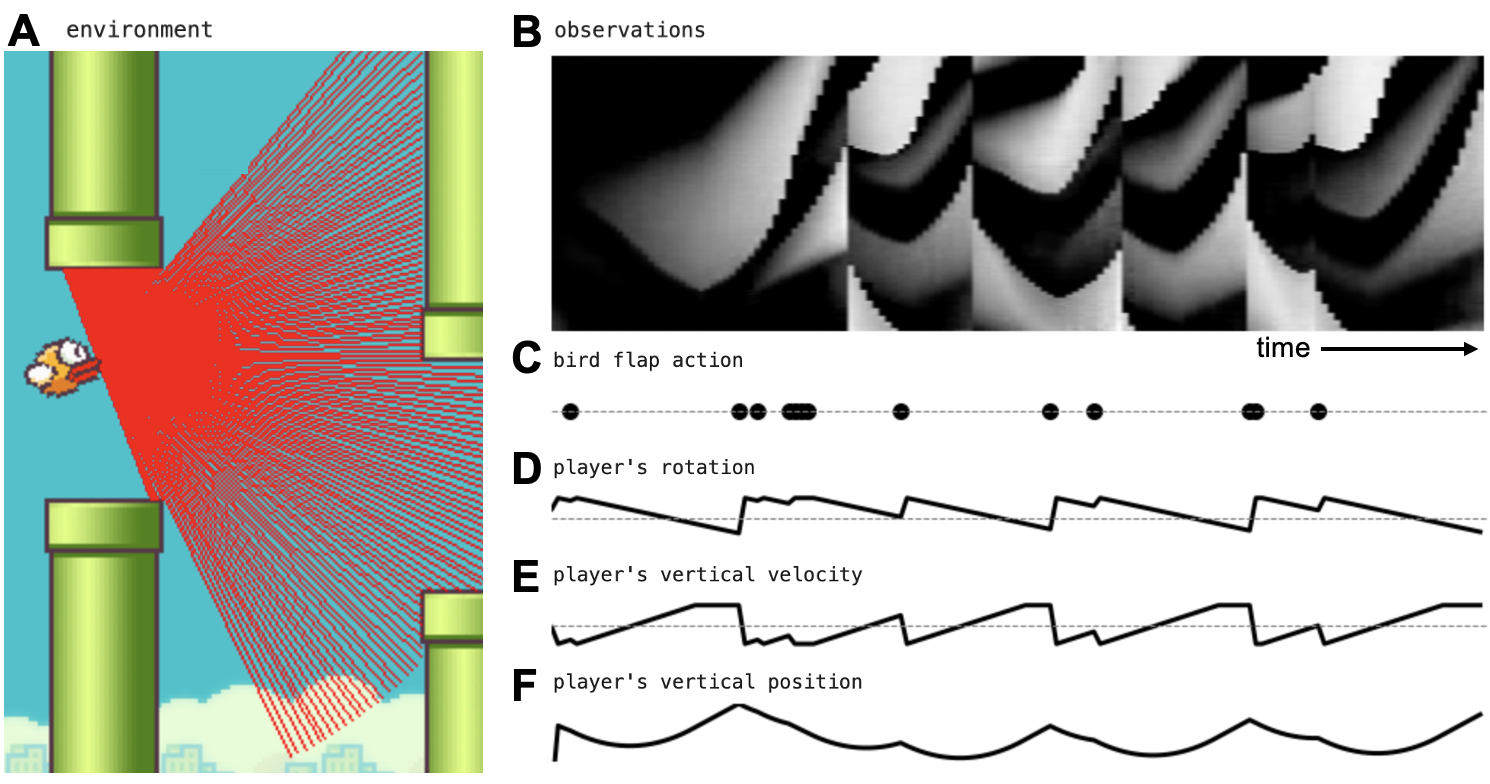}
  \caption{Flappy Bird environment with lidar. (A) The environment. (B) The agent only observes the lidar signal as a function of time. (C) The only available actions are no-op and flap. (D-F) The environment provides true variables of the world, such as the player's rotation angle, vertical velocity, and position.}
  \label{fig:env}
\end{figure}

\subsection{Environment and Task Description}
We use an experimental environment based on the Flappy Bird game\footnote{The original implementation of the environment can be found at https://github.com/markub3327/flappy-bird-gymnasium.} (Figure~\ref{fig:env}A), where the bird must successfully pass through layers of pipes by performing either a no-op action or a flap action, which propels the bird upward (Figure~\ref{fig:env}C). The agent receives only LIDAR measurements as observations, represented by a 180-dimensional vector indicating the distance to the nearest obstacle for each ray angle (Figure~\ref{fig:env}B). Separately, the environment also provides the true underlying world features, such as the bird's rotation, vertical velocity, and position (Figure~\ref{fig:env}D-F), as well as information about the pipes.

The agent (bird) receives reward every time it successfully passes through a pair of pipes. The episode ends whenever the bird touches any obstacle (pipe, floor, or ceiling). However, in this paper we focus exclusively on the world model aspect of this environment, and so the rewards are not considered.

For the experiments in this paper, we use 10,000 training rollouts of the environment collected by an expert DQN policy (which turns into a random policy with probability 1\% every timestep; this randomness was included to encourage off-policy exploration). For testing, we use 2,000 rollouts of the DQN policy and 2,000 rollouts from a random policy (which chooses action "flap" with probability 7.5\% every timestep).

\subsection{Network Architecture and Training}
\begin{figure}[h]
  \centering
  \includegraphics[width=.8\textwidth]{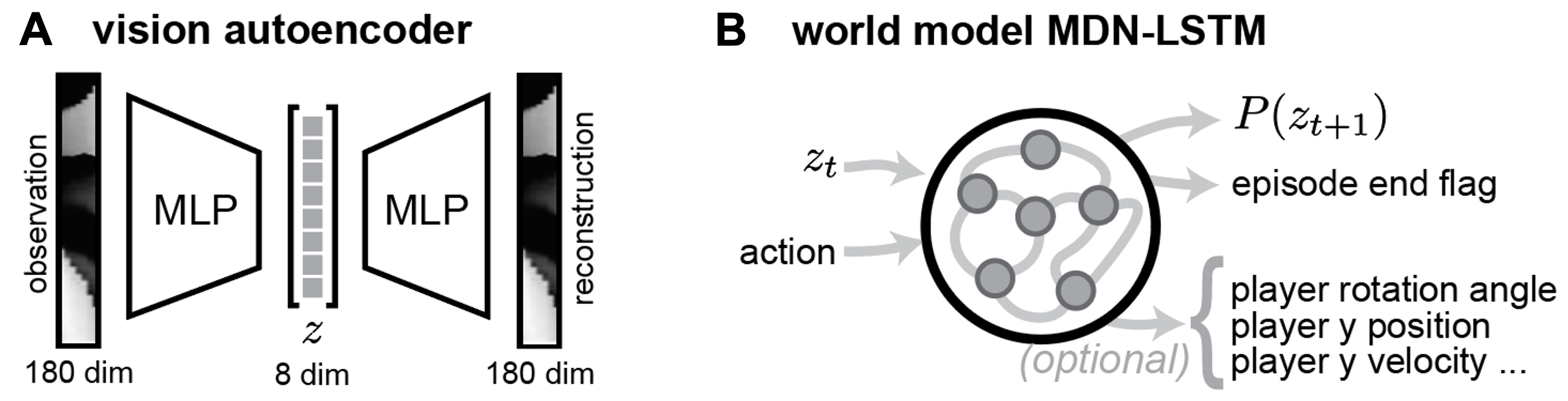}
  \caption{Network architecture and training setup. (A) The vision autoencoder compresses the 180-dimensional raw observations into an 8-dimensional latent space. (B) The world model MDN-LSTM takes the current latent observation vector and action as inputs, and predicts the distribution of the next latent vector and an episode end flag. Optionally, the LSTM's hidden state is encouraged to capture true world variables (such as player rotation angle, y position, y velocity, etc.) through a linear probe.}
  \label{fig:arch}
\end{figure}
The network architecture is based on that of the world models in the paper by \cite{NEURIPS2018_2de5d166}.

First, raw 180-dimensional observations are transformed into an 8-dimensional latent space using an autoencoder (Figure~\ref{fig:arch}A). The autoencoder consists of an encoder multi-layer perceptron (MLP) $E$ that maps the high-dimensional observations to the lower 8-dimensional latent space (outputting the mean and variance of an 8-dimensional factored gaussian distribution), and a decoder MLP $D$ that reconstructs the original observations from the latent representation.

The autoencoder is trained on the collected rollouts, with L2 regularization, with the following loss:
\begin{equation}
    L_{AE} = \sum_i(\hat x_i - x_i)^2+\frac 1 2 \sum_i \lVert E(x_i)\rVert^2,
\end{equation}
where 
\begin{align}
    \hat x_i = D(z_i); \hspace{2mm} z_i \sim \mathcal N(\mu_i, \sigma_i^2); \hspace{2mm} E(x_i)=(\mu_i, \sigma_i).
\end{align}

After training, the weights of the autoencoder are frozen. 

The world model is a long short-term memory RNN (LSTM; Hochreiter \& Schmidhuber (1997); Figure~\ref{fig:arch}B). This RNN takes as input the current latent observation vector ($z_t$) and action $a_t$, and outputs the distribution of the next latent vector ($P(\tilde{z}_{t+1})$) and an episode end flag $\tilde{e}_{t+1}$. The distribution of latent vectors is represented using a mixture of 5 Gaussian variables (following the implementation of the mixture density network; \cite{graves2014generatingsequencesrecurrentneural, bishop1994mdn}) -- that is, the network outputs 5 tuples $(\mu^{(i)}, \sigma^{(i)})$ to represent the distribution $P(\tilde{z}_{t+1})$.

The loss function is
\begin{equation}
    L_{pred} = -\frac 1 n\sum_t \log \left[\frac{1}{5}\sum_{i=1}^5 P(z_{t+1} \mid \mu^{(i)}_t, \sigma^{(i)}_t)\right]+\frac{10} n\sum_t (e_t-\tilde{e}_t)^2,
\end{equation}
where $n$ is the number of timesteps in an episode.

Our key contribution is the (optional) addition of linear probes to the network, which attempt to decode the true underlying world features from the LSTM's hidden state. The loss function is modified to include a term proportional to the mean squared error of the linear probes, with a hyperparameter multiplier $\lambda$. The total loss is the sum of the predictive loss and the linear probe mean squared error:
\begin{align}
    L_{total} &= L_{pred} + \lambda \cdot MSE(\text{linear probe})\\
    &=L_{pred} + \lambda \sum_{k,t}\left(W_kr_t-f_{k,t}\right)^2,
\end{align}
where $W_k$ are the linear weights for world feature $k$; $r_t$ are the features in the recurrent layer of the LSRM at time $t$, and $f_{k,t}$ is the value of the world feature $k$ for the timestep $t$.

\section{Results}

\subsection{Improved Next State Prediction Loss}
To assess the effects of adding linear probes during training, we trained world model RNNs for various values of $\lambda$ (20 random seeds for each choice of $\lambda$). If $\lambda=0$, the setting is equivalent to training on the predictive loss only, without any linear probe component. 

\begin{figure}[h]
  \centering
  \includegraphics[width=1.0\textwidth]{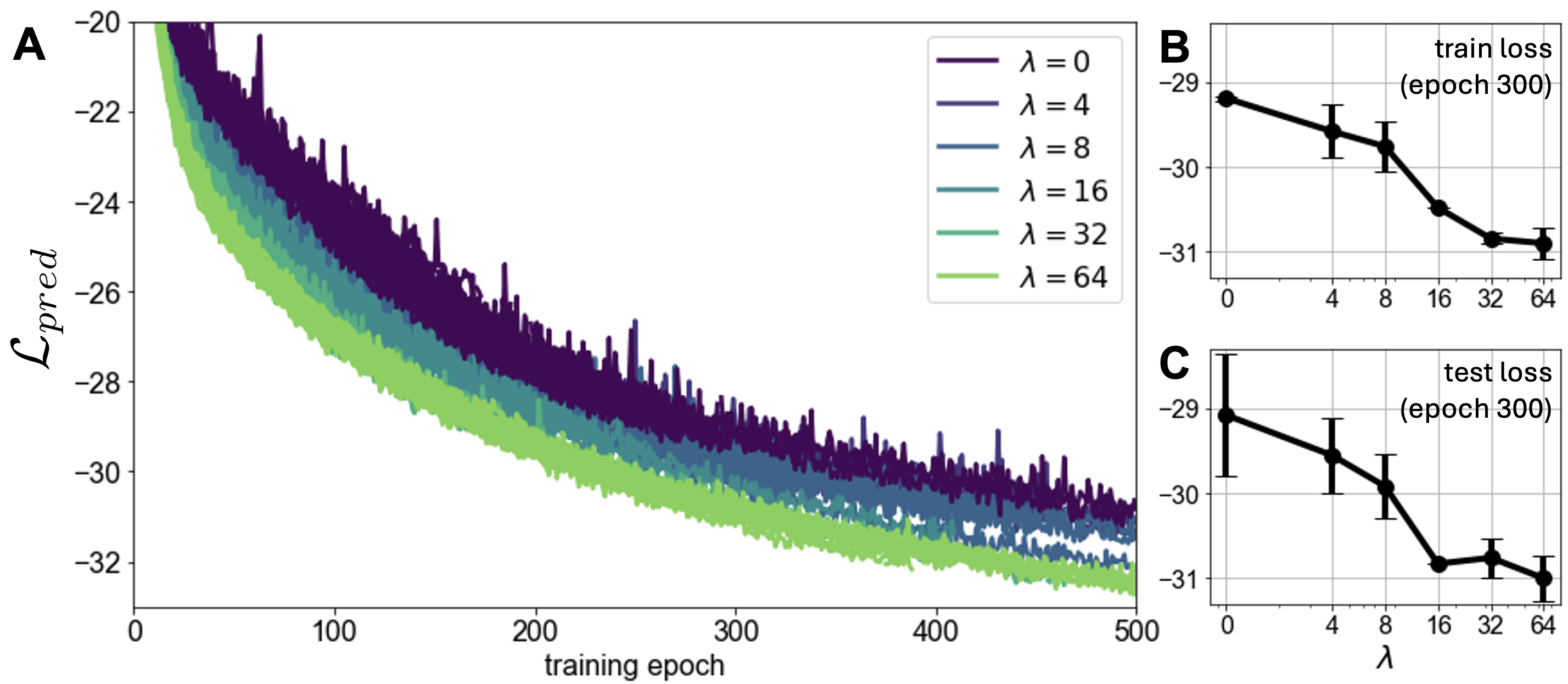}
  \caption{The effect of increasing the linear probe weight $\lambda$ on the original (next latent state prediction) loss. Both training (A, B) and test (C) predictive losses decrease as $\lambda$ increases. For every choice of $\lambda$, 20 RNNs were initialized from different random seeds. Error bars in panels (B) and (C) indicate the s.d.}
  \label{fig:loss}
\end{figure}

Perhaps paradoxically, increasing the hyperparameter $\lambda$, which put more weight on the linear probe component of the loss, did not hurt the predictive loss component. Instead, both the training and test predictive losses decreased as $\lambda$ was increased (Figure~\ref{fig:loss}A-C). In fact, the predictive loss as a function of $\lambda$ continued to decrease through values as high as $\lambda=64$. 

This result suggested that the addition of the linear probe component to the loss function improved the network's performance without compromising its predictive capabilities. For the rest of this paper, only networks with $\lambda=0$ and $\lambda=64$ will be considered and compared together.

\subsection{Improved Decodability of World Features}
To assess the sophistication of the world model, we examined the performance of linear probes in decoding true world variables from the network's hidden state. 

\begin{figure}[h]
  \centering
  \includegraphics[width=1.0\textwidth]{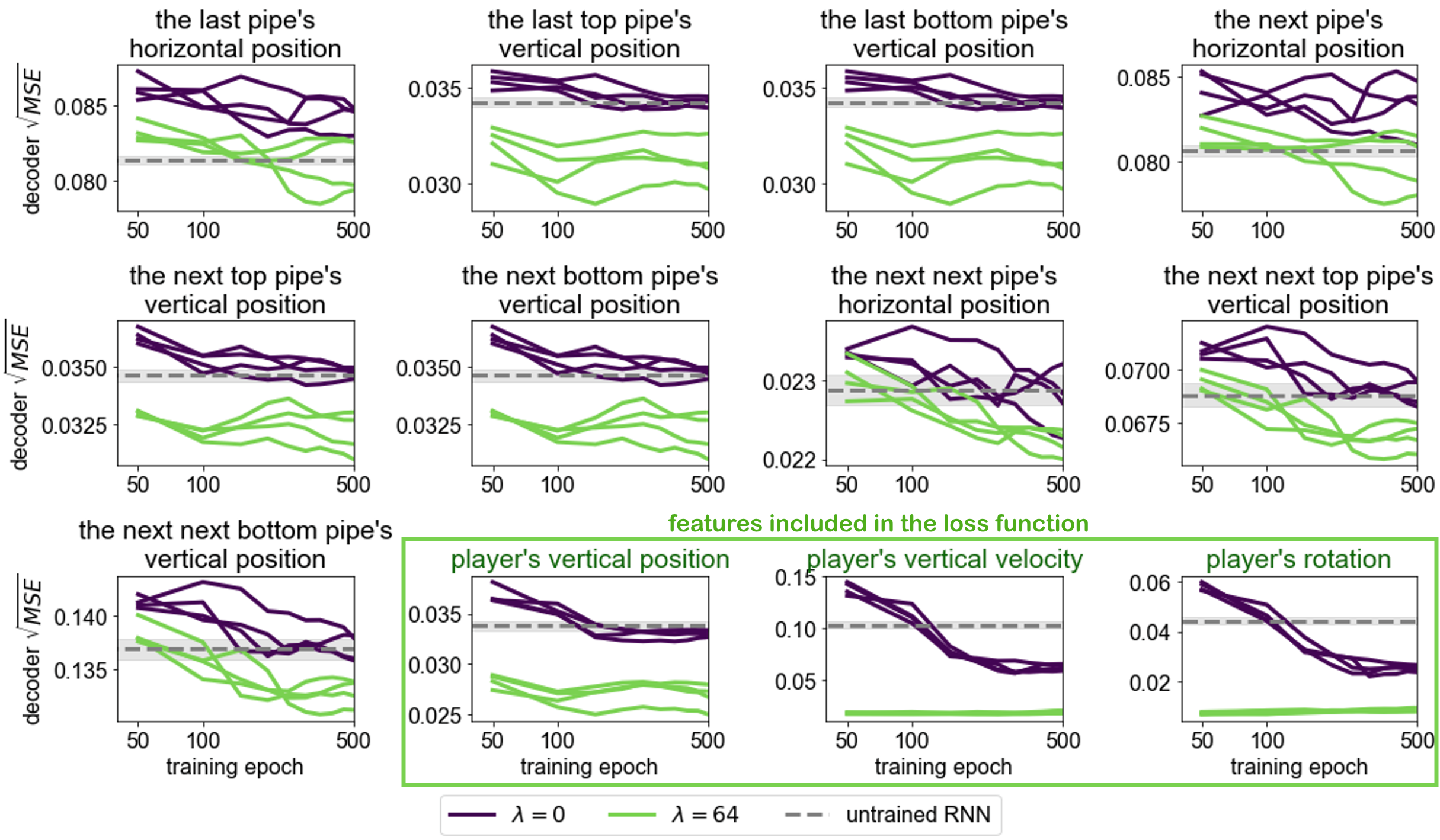}
  \caption{Decodability of world features from the network's hidden state for $\lambda=0$ and $\lambda=64$. Every panel represents one world feature. The last three panels (player's vertical position, velocity, and rotation; green) represent features which were explicitly included in the loss function for $\lambda=64$. Note that even the features which were not explicitly part of the losss function exhibit decodability above that of an untrained randomly-initialized network, but only for $\lambda=64$. For every choice of $\lambda$, four RNNs  initialized from different random seeds are shown.}
  \label{fig:decode}
\end{figure}

As expected, the world features that were included in the linear probe loss during training are more easily decodable for $\lambda=64$ than for $\lambda=0$ (Figure~\ref{fig:decode}; last three panels -- player's vertical position, velocity, and rotation angle). 

Interestingly however, only for $\lambda=64$, even features which were not explicitly part of the loss function exhibited decodability above that of an untrained randomly-initialized network (Figure~\ref{fig:decode}; all panels). This finding indicated that the addition of the linear probe component encouraged the network to develop a more comprehensive representation of the underlying world features, perhaps developing a more sophisticated world model.

\subsection{Reduced Distribution Drift}
Next, we tested the network's ability to predict how the latent state changes over time, without receiving constant feedback input from the environment. Specifically, we chose a certain timestep, after which the network was cut off from the environment (stopped receiving the true input latent vector every timestep). Instead, after the cutoff timestep, we sampled the next predicted latent state from the network's output and fed it back as its input in the subsequent step. 

\begin{figure}[h]
  \centering
  \includegraphics[width=0.9\textwidth]{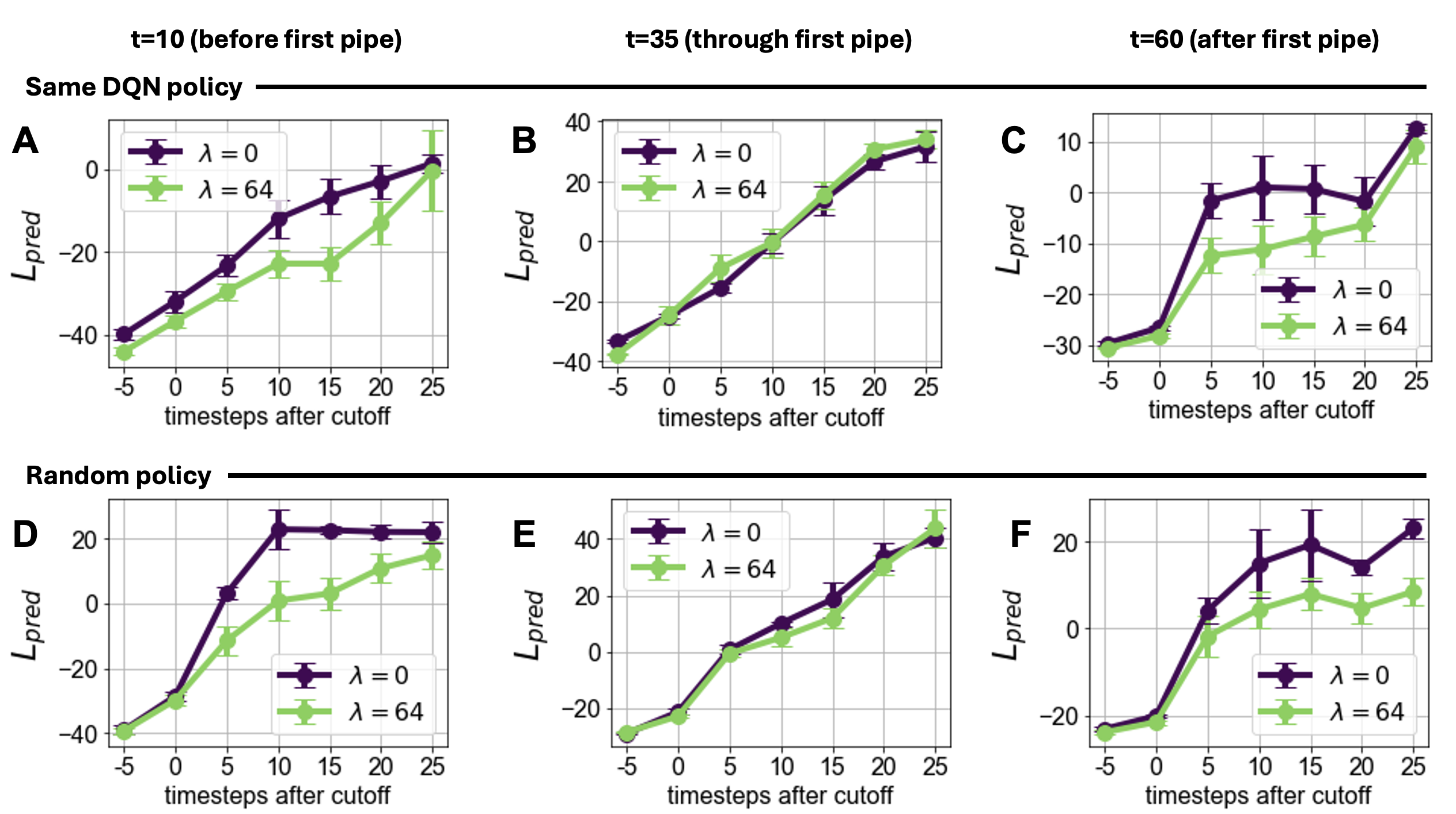}
  \caption{Distribution drift comparison. Networks trained with the linear probe ($\lambda = 64$) exhibit reduced distribution drift compared to those without the probe ($\lambda = 0$) for both the training policy and a random policy. However, note that this result does not hold for the timesteps near t=35, which correspond to the bird going through the first pair of pipes. The error bars represent s.d. across 10 random seeds.}
  \label{fig:drift}
\end{figure}

Figure \ref{fig:drift} compares the distribution drift between networks trained with and without the linear probe. Networks trained with the linear probe ($\lambda = 64$) exhibited smaller growth in the loss of predicting the real latent state, indicating reduced distribution drift, but only for timesteps when the bird was flying between successive pipe encounters (t=10 and t=60; Figure~\ref{fig:drift}A, C). Interestingly, during the timesteps when the bird was flying through a pair of pipes, this result did not hold. This may have to do with the fact that the changes in game state are much less predictable in those timesteps.

The results were qualitatively equivalent for both the training policy (DQN) and a random policy (Figure~\ref{fig:drift}D-F). This finding suggests effectiveness of the linear probes in maintaining a more accurate prediction of the latent observation state over time.

\subsection{Scaling Properties}
\begin{figure}[h]
  \centering
  \includegraphics[width=.8\textwidth]{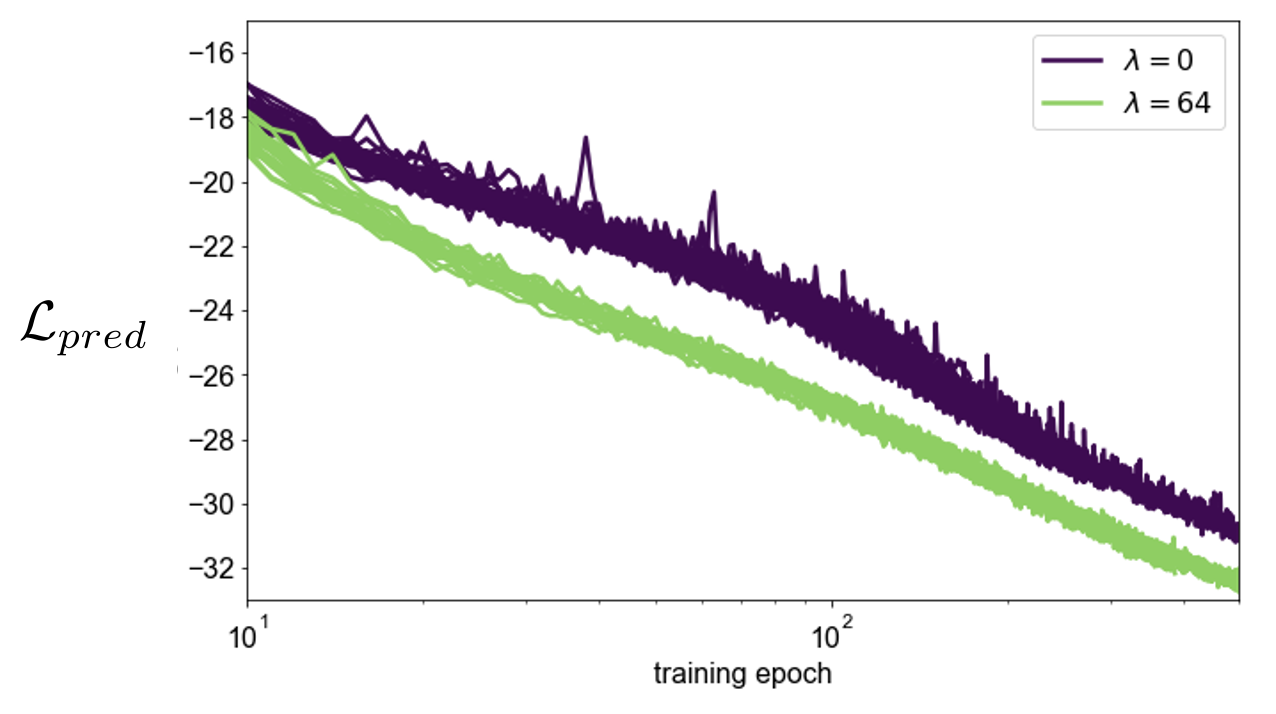}
  \caption{Scaling law of predictive loss with respect to training time. The predictive loss ($L_{pred}$) is plotted against the training epoch on a log scale for two different values of the linear probe weight: $\lambda=0$ (purple) and $\lambda=64$ (green). This plot highlights that while the linear probe improves the network's performance, it does not increase the slope of the scaling law of predictive loss with training time. Training curves of 20 randomly-initialized networks is plotted for every choice of $\lambda$.}
  \label{fig:scaling_one}
\end{figure}

The predictive component of the loss followed a scaling law with respect to training time (Figure~\ref{fig:scaling_one}), for both $\lambda=0$ and $\lambda=64$. However, the scaling law for $\lambda=64$ was smoother, as well as shifted down, compared to the one for networks trained with $\lambda=0$.

\begin{figure}[h]
  \centering
  \includegraphics[width=.8\textwidth]{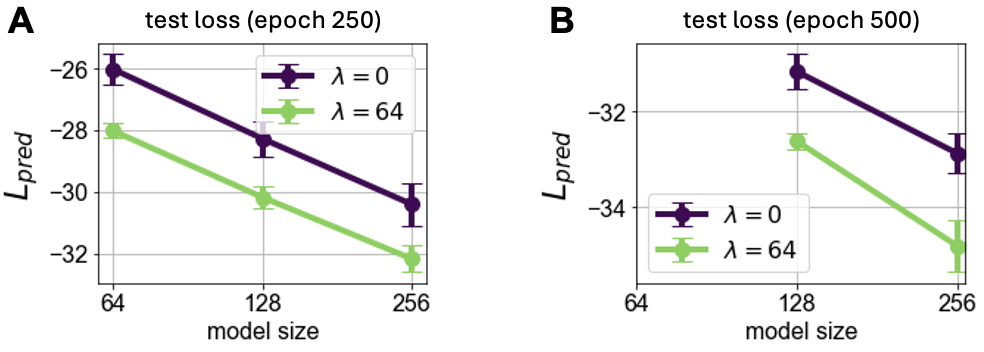}
  \caption{Scaling laws for model size. The performance of networks with and without the linear probe follows a scaling law, but the curve for networks trained with $\lambda = 64$ is shifted down, indicating better performance for a given model size. This holds true at both (A) epoch 250 and (B) epoch 500.}
  \label{fig:scaling}
\end{figure}

The loss of networks with and without the linear probe also followed a scaling law with respect to model size (Figure~\ref{fig:scaling}). However, the curve for networks trained with $\lambda = 64$ was shifted down, such that a network of a given size trained with the linear probe achieves performance roughly equivalent to a network twice the size trained without the probe. This result held true at both epoch 250 and 500. 

This finding suggested that the linear probe component provides a consistent performance benefit across different stages of training and different model sizes.

\subsection{Training Stability}
\begin{figure}[h]
  \centering
  \includegraphics[width=.6\textwidth]{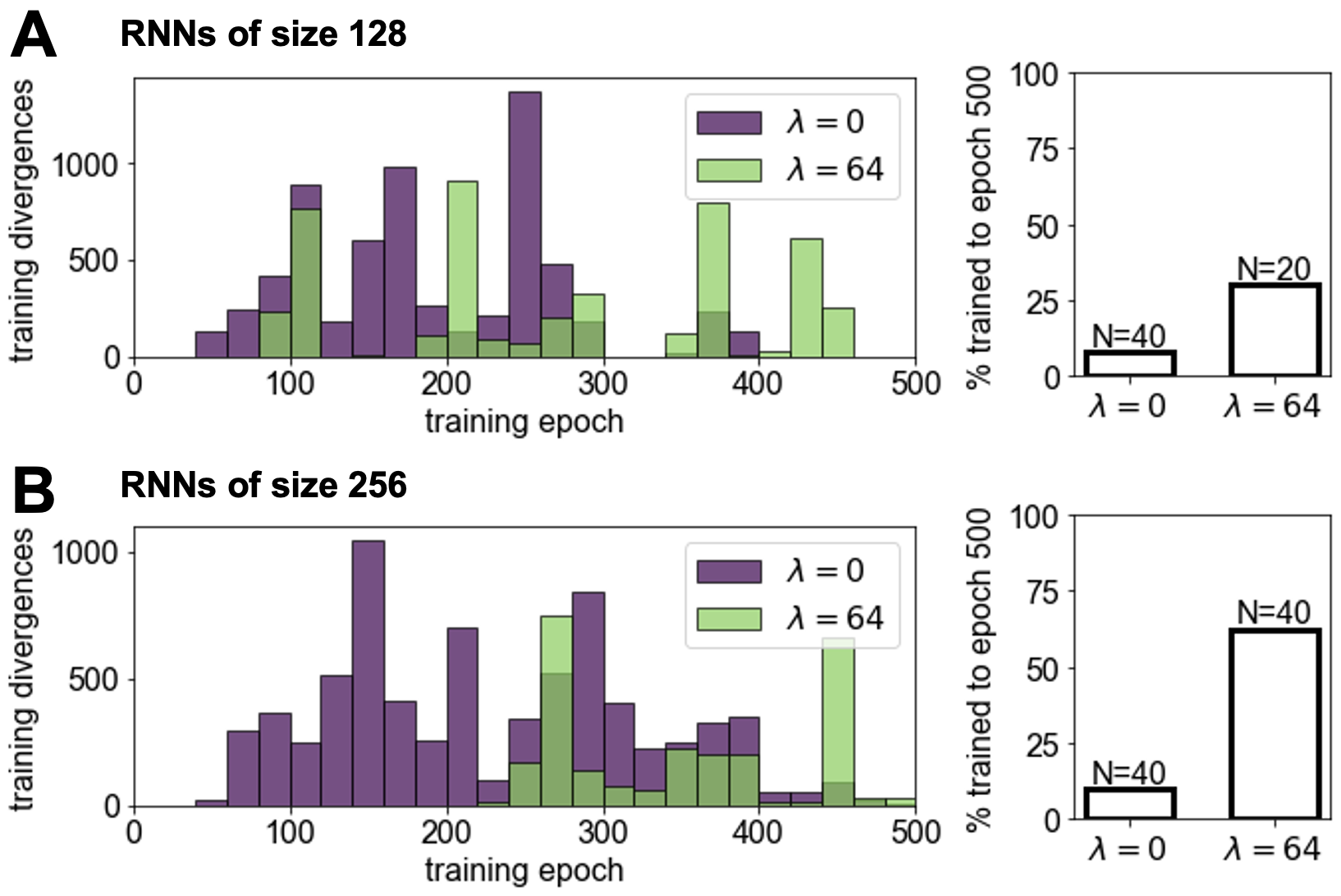}
  \caption{Training stability with and without the linear probe. Networks trained with the linear probe are more likely to reach epoch 500 without diverging. Here, results are shown for RNNs of hidden size 128 (A) and 256 (B).}
  \label{fig:stability}
\end{figure}

Finally, adding the linear probe mean squared error component to the loss function improved training stability (Figure~\ref{fig:stability}). It resulted in fewer instances of exploding gradients and training divergences, with the divergences additionally happening later in training for $\lambda=64$ than for $\lambda=0$. Networks trained with the linear probe are more likely to reach epoch 500 without diverging (30\% vs 7.5\% for RNNs of size 128, 62.5\% vs 10\% for RNNs of size 256; Figure~\ref{fig:stability}A-B, right).

\section{Discussion}
In this study, we sought to encourage the development of world models in a recurrent neural network trained end-to-end to predict the next observation of an agent. Specifically, we considered adding a term to the loss function of the network, which was proportional to the mean squared error of linear probes trying to decode the true world features from the hidden state of the model.

In our experiments, this addition demonstrated improved training and test loss, enhanced training stability, and more easily decodable world features. We also observed reduced distribution drift in certain cases for models trained with the linear probe. The scaling laws for both model size and training time maintained the same slope both with and without the additional of linear probe loss, consistent with findings in the literature on scaling laws for neural networks \citep{kaplan2020scalinglawsneurallanguage}. However, the networks trained with the linear probes had the loss curve shifted down, such that, for example, a network of a given size trained with the linear probe achieved performance roughly equivalent to a network twice the size trained without the probe.
This technique may prove especially advantageous in compute-limited settings or when aiming to achieve the best performance with smaller models, for example when deploying models on-site in medical imaging settings \citep{zhou2017deepsupervisionpancreaticcyst, lin2021automated, mishra2019ultrasound, LEI2018290}.

Our findings support the annealing technique, wherein the linear probe loss term is applied only during the initial stages of training and gradually diminished to zero. The analysis of the loss curve (Figure~\ref{fig:scaling_one}) shows that the slope stabilizes after a certain number of steps, indicating that the linear probes have the most pronounced impact during the early stages of training when the network is actively learning its foundational representations. This suggests that the technique might be particularly beneficial at the beginning of training and less critical in later stages. By focusing the influence of the probes early on, this approach enhances the decodability of world features without imposing a lasting computational burden. Furthermore, transitioning to normal training procedures after the initial phase would remove the additional data requirement, making this technique both efficient and scalable. Testing this phased application of the technique is a promising direction for future work.

Moreover, the concept of leveraging additional sensory inputs aligns with applications in robotic systems \citep{pmlr-v205-wu23c}. When a robot is augmented with inexpensive, additional sensors, decoding these inputs through the network's latent representations can enhance its ability to interpret and model the environment. This process facilitates improved learning by enriching the internal state of the model with diverse sensory information. This technique is thus promising for scalable and cost-effective advances in robotics, where this addition of sensors can unlock significant performance gains, while allowing the models to be smaller and therefore run faster \citep{firoozi2024foundation}.

Overall, this study contributes to our understanding of how to encourage the development of robust world models in end-to-end trained predictive networks. Further research in this direction may yield valuable insight for advancing the field of artificial intelligence and developing more capable agents.

\subsubsection*{Acknowledgments}
A.Z. would like to express gratitude to Prof. Phillip Isola, William Shen, and Prof. Ila Fiete at MIT for their helpful guidance, feedback, and inspiration throughout this project.

\bibliography{Andrii_references}
\bibliographystyle{iclr2025_conference}

\end{document}